\documentclass[runningheads]{llncs}
\usepackage{graphicx}
\usepackage{amsmath,amssymb} 
\usepackage{color}
\usepackage[width=122mm,left=12mm,paperwidth=146mm,height=193mm,top=12mm,paperheight=217mm]{geometry}

\usepackage[ruled]{algorithm2e}
\usepackage[bookmarks=false]{hyperref}
\usepackage{cite}

\graphicspath{ {images/} }

\begin{document}
\pagestyle{headings}
\mainmatter

\title{Multi-Class Multi-Object Tracking using Changing Point Detection} 

\titlerunning{Multi-Class Multi-Object Tracking using Changing Point Detection}

\authorrunning{B. Lee et al.}

\author{Byungjae Lee$^1$, Enkhbayar Erdenee$^1$, Songguo Jin$^1$, and Phill Kyu Rhee$^1$}


\institute{Inha University$^1$
}

\maketitle

\begin{abstract}
This paper presents a robust multi-class multi-object tracking (MCMOT) formulated by a Bayesian filtering framework. Multi-object tracking for unlimited object classes is conducted by combining detection responses and changing  point detection (CPD) algorithm. The CPD model is used to observe abrupt or abnormal changes due to a drift and an occlusion based spatiotemporal characteristics of track states. The ensemble of convolutional neural network (CNN) based object detector and Lucas-Kanede Tracker (KLT) based motion detector is employed to compute the likelihoods of foreground regions as the detection responses of different object classes. Extensive experiments are performed using lately introduced challenging benchmark videos; ImageNet VID and MOT benchmark dataset. The comparison to state-of-the-art video tracking techniques shows very encouraging results. 

\end{abstract}

\section{Introduction}

Multi-object tracking (MOT) is emerging technology employed in many real-world applications such as video security, gesture recognition, robot vision, and human robot interaction [1-15]. The challenge is drifts of tracking points due to appearance variations caused by noises, illumination, pose, cluttered background, interactions, occlusion, and camera movement. Most MOT methods are suffered from varying numbers of objects, and leading to performance degradation and tracking accuracy impairments in cluttered backgrounds. However, most of them only focus on a limited categories, usually people or vehicle tracking. MOT with unlimited classes of objects has been rarely studied due to very complex and high computation requirements.

The Bayesian filter consists of the motion dynamics and observation models which estimates posterior likelihoods. One of the Bayesian filter based object tracking methods is Markov chain Monte Carlo (MCMC)-based method \cite{Ref2,Ref3,Ref4,Ref5}, which can handle various object moves and interactions of multiple objects. Most MCMC based methods assume that the number of objects would not change over time, which is not acceptable in a real world applications. Reversible jump MCMC (RJMCMC) was proposed by \cite{Ref2,Ref4}, where a variable number of objects with different motion changes, such as update, swap, birth, and death moves. They start a new track by initializing a new object or terminates currently tracked object by eliminating the object.

Even MCMC based MOT approaches were successful to some extent, computational overheads are very high due to a high-dimensional state space. The variations in appearances, the interaction and occlusions and changing number of moving objects are challenging, which require high computation overheads. Saka et. al. \cite{Ref1} proposes a MCMC sampling with low computation overhead by separating motion dynamics into birth and death moves and the iteration loop of the Markov chain for motion moves of update and swap. If the moves of birth and death are determined inside of the MCMC chain, it requires the dimension changes in the MCMC sampling approaches as \cite{Ref2,Ref3}. Since the Markov chain has no dimension variation in the iteration loop by separating the moves of birth and death, it can reach to stationary states with less computation overhead \cite{Ref1,Ref6}. However, such a simple approach for separating birth and death dynamics cannot deal with complex situations that occur in MOT. Many of them are suffered from track drifts due to appearance variations.

In this paper, we propose a robust multi-class multi-object tracking (MCMOT) that conducts unlimited object classes by combining detection responses and changing  point detection (CPD) algorithm. With advances of deep learning based object detection technology such as Faster R-CNN \cite{Ref28}, and ResNet \cite{Ref29}, it becomes feasible to adopt a detector ensemble with unlimited classes of objects. The detector ensemble combines the model based detector implemented by Faster R-CNN \cite{Ref28} and the motion detector by Lucas-Kanade Tracker (KLT) algorithm \cite{Ref26}. The method separates the motion dynamic model of Bayesian filter into the entity transitions and motion moves. The entity transitions are modeled as the birth and death events. Observation likelihood is calculated by more sophisticated deep learning based data-driven algorithm. Drift problem which is one of the most cumbersome problems in object tracking is attacked by a CPD algorithm similarly to \cite{Ref24}. Assuming the smoothness of motion dynamics, the abrupt changes of the observation are dealt with the CPD algorithm, whereas the abrupt changes are associated illuminations, cluttered backgrounds, poses, and scales. The main contributions of the paper are below:

\begin{itemize}
    \item[$\bullet$] MCMOT can track varying number of objects with unlimited classes which is formulated as a way to estimate a likelihood of foreground regions with optimal smoothness. Departing from the likelihood estimation only belong to limited type of objects, such as pedestrian or vehicles, efficient convolutional neural network (CNN) based multi-class object detector is employed to compute the likelihoods of multiple object classes.
    \item[$\bullet$] Changing point detection is proposed for a tracking failure assessment by exploiting static observations as well as dynamic ones. Drifts in MCMOT are investigated by detecting such abrupt change points between stationary time series that represent track segment.
\end{itemize}

This paper is organized as follows. We review related work in Section 2. In Section 3, the outline of MCMOT is discussed. Section 4 introduces our proposed tracking method. Section 5 describes the experiments, and concluding remarks and future directions are discussed in Section 6.

\section{Related Work}

\subsection{Multi Object Tracking}

Recent research in MOT has focused on the tracking-by-detection principal to perform data association based on linking object detections through a video sequence. Majority of the batch methods formulates MOT with future frame's information to get better data association via hierarchical tracks association \cite{Ref13}, network flows \cite{Ref12}, and global trajectory optimization \cite{Ref11}. However, batch methods have higher computational cost relatively. Whereas online methods only consider past and current frame's information to solve the data association problem. Online methods are more suitable for real-time application, but those are likely to drift since objects in a video show significant variations in appearances due to noises, illuminations, poses, viewing angles, occlusions, and shadows, some objects enters or leaves the scene, and sometimes show sharp turns and abrupt stops. Dynamically varying number of objects is difficult to handle, especially when track crowded or high traffic objects in \cite{Ref9,Ref10,Ref14}. Most MOT methods relying on the observation of different features are prone to result in drifts. Against this nonstationarity and nonlinearity, stochastic-based tracking \cite{Ref22,Ref23,Ref24} appear superior to deterministic based tracking such as Kalman filter \cite{Ref33} or particle filter \cite{Ref2}.

\subsection{Convolutional Neural Network}

In the last few years, considerable improvements have been appeared in the computer vision task using CNN. One of the particularly remarkable studies is R-CNN \cite{Ref34}. They transferred CNN based image classification task to CNN based object detection task using region-based approach. SPPnet \cite{Ref35} and Fast R-CNN \cite{Ref36} extend R-CNN by pooling convolutional features from a shared convolutional feature map. More recently, RPN \cite{Ref28} is suggested to generate region proposals within R-CNN framework using RPN. Those region-based CNN pipelines outperform all the previous works by a significant margin. Despite such great success of CNNs, only a few number of MOT algorithms using the representations from CNNs have been proposed \cite{Ref20,Ref21,Ref22}. In \cite{Ref20,Ref21}, they proposed a CNN based framework with simple object tracking algorithm for MOT task in ImageNet VID. In \cite{Ref22}, they used CNN based object detector for MOT Challenge \cite{Ref32}. Our experiment adopts this paradigm of region based CNN to build observation model.

\begin{figure*}
\centering
  \includegraphics[width=0.9\textwidth]{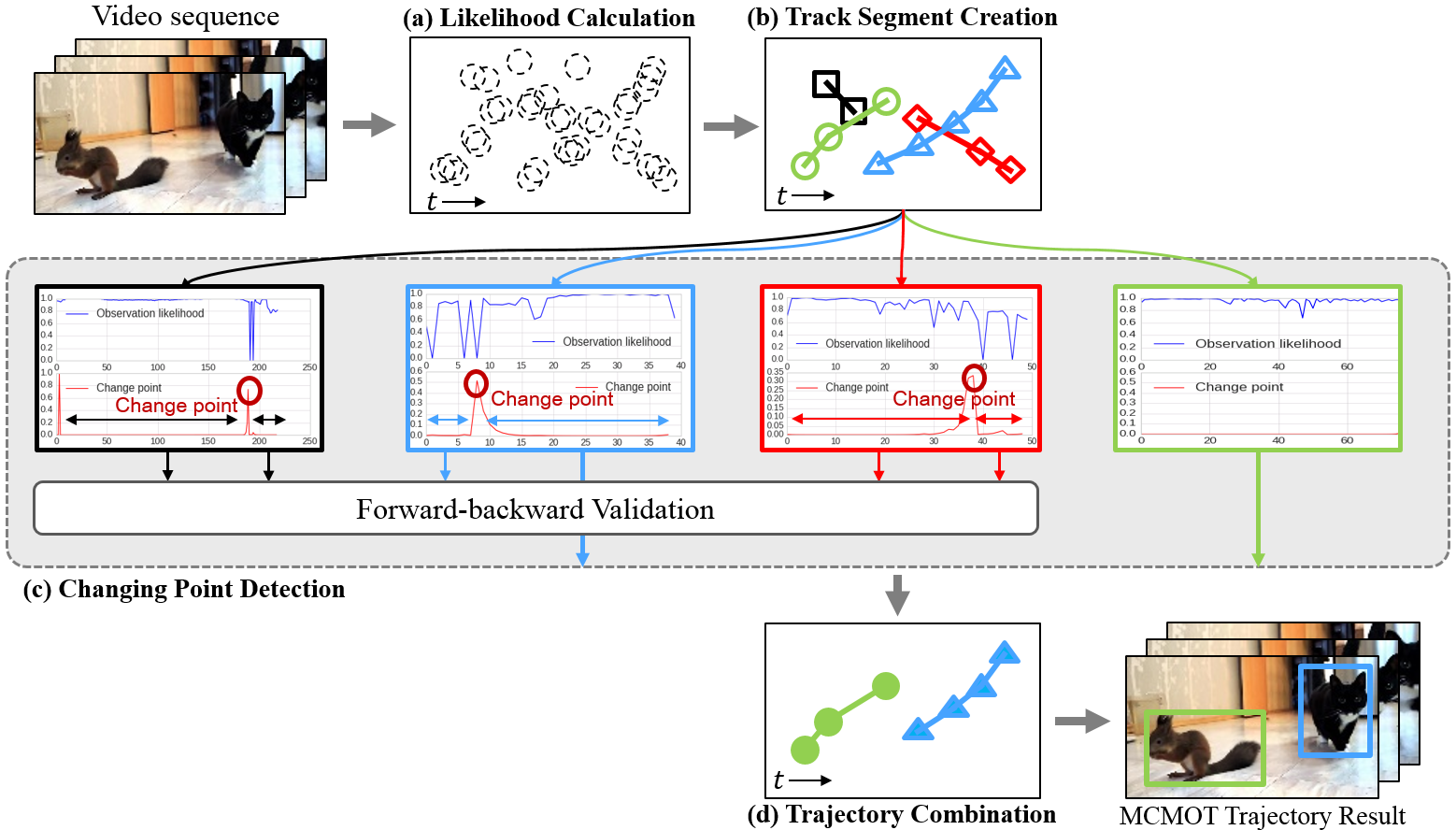}
\caption{MCMOT framework has four major steps: (a) Likelihood calculation based on observation models, (b) Track segment creation, (c) Changing point detection, and (d) Trajectory combination. The drifts in segments are effectively controlled by changing point detection algorithm with forward-backward validation.}
\label{fig:1}
\end{figure*}

\section{The Outline of MCMOT}

We propose an efficient multi-class multi-object tracker, called MCMOT that can deal with object birth, death, occlusion, interaction, and drift efficiently. MCMOT may fail due to the miscalculations of the observation likelihood, interaction model, entry model, and motion model. The objective of MCMOT is to stop the tracking as quick as possible if a drift occurs, recover from the wrong decisions, and to continue tracking. Fig.~\ref{fig:1} illustrates the main concept of our framework.

In MCMOT, objects are denoted by bounding boxes which are tracked by a tracking algorithm. In the tracking algorithm, if a possible interaction or occlusion is detected, the trajectory is split into several parts, called track segments. The combination of track segments is controlled by CPD. Considering fallible decision tracker points, CPD monitors a drift due to abnormal events, abrupt changing environments by comparing the localized bounding boxes by the observations within the segment. The motion-based tracking component facilitates KLT \cite{Ref26} adaptive for predicting the region of a next tracking point. The model-based component consists of the global object detector and adaptive local detector. We use a deep feature based multi-class object detector \cite{Ref28} as the global and local object detector. One can notice that the number of object categories can be readily extended depending on object detector capability.

\section{Multi-Class Multi-Object Tracking}

MCMOT employs an data-driven approach which investigates the events caused by object-level events, object birth and death, inter-object level events, i.e., interaction and occlusion between objects, and tracking level events, e.g. track birth, update, and death. Possible drifts due to the observation failures are dealt with the abnormality detection method based on the changing point detection. 

We define track segments using the birth and death detection. Only visible objects are tracked, the holistic trajectory divided into several track segments, if an occlusion happens as in \cite{Ref16}. If the object becomes ambiguous due to occlusion or noise, the track segment is terminated (associated object death), and the tracker will restart tracking (associated object birth) nearby the terminated tracking point if the same object reoccurs, and the track segment is continuously built, if it is required, or a new track segment is started and merged later. 

\subsection{Observation Model}

We define observation model (observation likelihood) $P{\rm (}{\bf z}_{t} |{\bf x}_{t} {\rm )}$ in this section. The observation likelihood for tracked objects need to estimate both the object class and accurate location. MCMOT ensembles object detectors with different characteristics to calculate the observation likelihood accurately. Since the dimensionality of the scene state is allowed to be varied, the measure is defined as the ratio of the likelihoods of the existence and non-existence. As the likelihood of the non-existence set cannot be measured, we adopt a soft max $f{\rm (}\cdot {\rm )}$ of the likelihood model, as in \cite{Ref18}. 
\begin{equation} \label{GrindEQ__1_} 
\frac{P{\rm (}\tilde{{\bf o}}_{t} |{\bf o}_{id,t} {\rm )}}{P{\rm (}\tilde{{\bf o}}_{t} |\not {\bf o}_{id,t} {\rm )}} =\exp \left(\sum _{e}f{\rm (}\lambda _{e} \log _{e} {\rm (}\tilde{{\bf o}}_{t} |{\bf o}_{id,t} {\rm )} \right) 
\end{equation} 
where $\not {\bf o}_{id,t} $ indicates the non-existence of object \textit{id}, $f$ soft max function, $\lambda _{e} $ the weight of object detector \textit{e}. The approach is expected to be robust to sporadic noises since each detector has its own pros and cons. We employ ensemble object detectors: deep feature based global object detector (GT), deep feature based local object detector (LT), color detector (CT), and motion detector (MT):

\begin{itemize}
    \item[$\bullet$] Global object detector (GT): Deep feature based object detector \cite{Ref28} in terms of hierarchical data model (HDM) \cite{HDM} is used.
    \item[$\bullet$] Local object detector (LT): By fine-tuning deep feature based object detector using confident track segments, issues due to false negatives can be minimized. Deep feature based object detector \cite{Ref28} is used for the local object detector.
    \item[$\bullet$] Color detector (CT): Similarity score between the observed appearance model and the reference target is calculated through Bhattacharyya distance \cite{Ref17} using RGB color histogram of the bounding box.
    \item[$\bullet$] Motion detector (MT): The presence of an object is checked by using KLT based motion detector \cite{Ref26} which detects the presence of motion in a scene. 
\end{itemize}

\subsection{Track Segment Creation}

The MCMOT models the tracking problem to determine optimal scene particles in a given video sequence. MCMOT can be thought as reallocation steps of objects from the current scene state to the next scene state repeatedly. First, the birth and death allocations are performed in the entity status transition step. Second, the intermediate track segments are built using the data-driven MCMC sampling step with the assumption that the appearances and positions of track segments change smoothly. In the final step, the detection of a track drift is conducted by a changing point detection algorithm to prevent possible drifts. Change point denotes a time step where the data attributes abruptly change \cite{Ref24} which is expected to be a drift starting point with high probability. We discuss the detail of the data-driven MCMC sampling, and entity status transition in follows.

\subsubsection{Date-Driven MCMC Sampling}

In a MCMC based sampling, the efficiency of the proposal density function is important since it affects much in constructing a Markov chain with stationary distribution, and thus affects much on tracking performance in practice. The proposal density function should be measurable and can be sampled efficiently from the proposal distribution \cite{Ref2}, which is proportional to a desired target distribution. We employ ``one object at a time'' strategy, whereas one object state is modified at a time, as in \cite{Ref2,Ref7}. Given a particle ${\bf x}_{t} $ at time t, the distribution of current proposal density function $\pi {\rm (}{\bf x'};{\rm x}_{t} {\rm )}$ is used to suggest for the next particle. In MCMOT, we assume that the distribution of the proposal density follows the pure motion model for the MCMC sampling, i.e., $\pi {\rm (}{\bf x'};{\rm x}_{t} {\rm )}\approx P{\rm (x}_{t+1} |{\bf x}_{t} {\rm )}$, as in \cite{Ref2}. Given a scene particle, i.e., a set of object states ${\bf x}_{t} $, a candidate scene particle ${\bf x}'_{t} $ is suggested by randomly selecting object ${\bf o}_{id,t} $, and then determines the proposed state ${\bf x'}_{t} $ relying the object ${\bf o}_{id,t} $ with  uniform probability assumption. In this paper, a strategy of data-driven proposal density \cite{Ref3} is employed to make the Markov chain has a better acceptance rate. MCMOT proposes a new state ${\bf o'}_{id,t} $ according to the informed proposal density with a mixture of the state moves to ensure motion smoothness as in \cite{Ref6}: 
\begin{equation} \label{GrindEQ__2_} 
\pi {\rm (}{\bf o}'_{id,t};{\bf x}_{t} {\bf )}=\left[\lambda _{1} \frac{1}{N} \sum _{s}p{\bf (o'}_{id,t} {\bf |o}_{id,t-1}^{{\bf (}s{\bf )}} {\bf \; )}+\lambda _{2} p{\bf (o}'_{id,t} {\bf |D}_{id,t} {\bf )} \right] 
\end{equation} 
where $\lambda _{1} +\lambda _{2} =1$. The first term is from the motion model and the second term from the detector ensemble and using the closest result from the all detection of object \textit{id}. 

Remind that the posterior probability for time-step \textit{t}-1 is assumed to be represented by a set of \textit{N} samples (scene particles). Given observations from the initial time to the current time \textit{t}, the calculation of the current posterior is done by MCMC sampling using \textit{N} samples. We use \textit{B} samples as burn-in samples \cite{Ref6}. \textit{B} burn-in samples are used initially and eliminated for the efficient convergence to a stationary state distribution. More details and other practical considerations about MCMC can be found in \cite{Ref42}. 

\subsubsection{Estimation of entity status transition}

The entity status is estimated by two binomial probabilities of the birth status and death status according to the entry model at time step \textit{t} and \textit{t}-1. Let $ES_{id,t}^{b} {\rm (}x,y{\rm )}=\nu \; {\rm (}\nu \in {\rm \{ 1,\; 0\} )}$ denote the birth status with $\nu $=1 indicating true, $\nu $= 0 false of an object \textit{id} in the potion${\rm (}x,y{\rm )}$. Similarly, $ES_{id,t}^{d} {\rm (}x,y{\rm )}=\nu \; $denotes death status. The posterior probability of entry status is defined at time \textit{t} as follows:

\begin{equation} \label{GrindEQ__3_} 
\resizebox{.93\hsize}{!}{$
P_{ES} {\rm (}{\bf o}_{id,t}^{} |{\bf o}_{id,t-1}^{} {\rm )}\approx \left\{\begin{array}{l} {P_{b} ={\rm \; }p{\rm (}ES_{id,t}^{b} {\rm (}x,y{\rm )}=1|{\bf o}_{id,1:t}^{} {\rm )\; ,\; \; if\; object\; }id{\rm \; exists\; time\; }t{\rm \; and\; not\; }t{\rm -1}} \\ {P_{d} =p{\rm (}ES_{id,t}^{d} {\rm (}x,y{\rm )}=1|{\bf o}_{id,1:t}^{} {\rm )\; ,\; \; if\; object\; }id{\rm \; exists\; at\; time\; }t{\rm -1\; and\; not\; }t} \\ {P_{a} =1-P_{d} ,{\rm \; \; \; if\; object\; }id{\rm \; exists\; at\; time\; }t{\rm -1\; and\; }t} \\ {P_{\emptyset } =1-P_{b} ,{\rm \; \; \; \; if\; object\; }id{\rm \; exists\; neither\; time\; }t{\rm \; nor\; }t{\rm -1}} \end{array}\right.  
$}
\end{equation} 

If a new object \textit{id} is observed by the observation likelihood mode at time \textit{t} in position (\textit{x},\textit{y}) which\textit{ }did not exist (detected) in time \textit{t}-1, the birth status of object \textit{id} $ES_{id,t}^{b} {\rm (}x,y{\rm )}$is set to 1, otherwise, it is set to 0. If an object \textit{id} is not observed by the detector ensemble at time \textit{t} in position (\textit{x},\textit{y}) which\textit{ }existed in time \textit{t}-1, the death status of object \textit{id}, i.e.,$ES_{id,t}^{b} {\rm (}x,y{\rm )}$is set to 1, otherwise, it is set to 0. 

\begin{figure}[t!]
\centering
\includegraphics[width=0.7\textwidth]{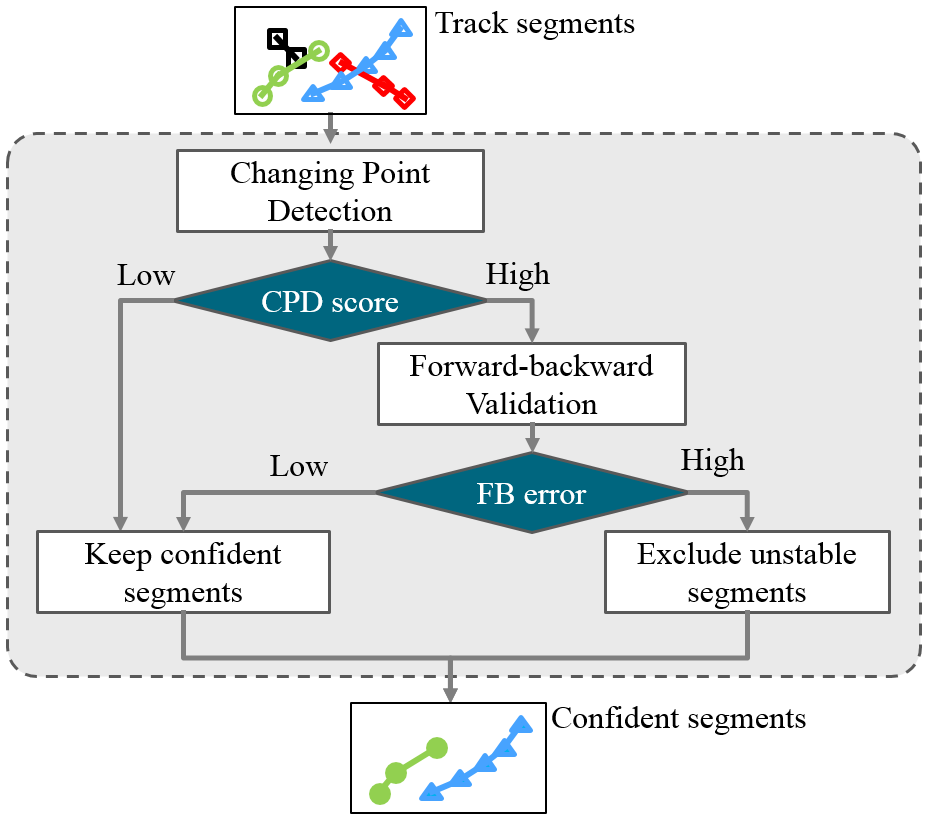}
\caption{Illustration of CPD. A change point score is calculated by the changing point detection algorithm. If the high change point score is detected, forward-backward error is checked from the detected change point. FB error checks whether the segment is drifted. A possible track drift is determined effectively by the change point detection method with forward-backward validation.}
\label{fig:2}
\end{figure}

\subsection{Changing Point Detection}

MCMOT may fail to track an object if it is occluded or confused by a cluttered background. MCMOT would determine whether or not a track is terminated or continues tracking. Drifts in MCMOT are investigated by detecting such abrupt change points between stationary time series that represent track segment. A higher response indicates a higher uncertainty with high possibility of a drift occurrence \cite{Ref25}. Two-stage time-series learning algorithm is used as in \cite{Ref24}, where a possible track drift is determined by a change point detection method \cite{Ref24} as follows. The 2${}^{nd}$ level time series is built using the scanned average responses to reduce outliers in the times series. The procedure to prevent drift is illustrated in Fig.~\ref{fig:2}. 

If high CPD response is detected on track segment, the forward-backward error (FB error) validation \cite{Ref7} is defined to estimate the confidence of a track segment by tracking in reverse sequence of the track segments. A given video, the confidence of track segment \textit{$\tau _{t}^{} $}is to be estimated. Let \textit{$\tau _{t}^{r} $} denotes the reverse sequential states, i.e., ${\bf o}_{id,t:1} ={\rm \{ }\hat{{\bf o}}_{id,t} ,\ldots ,\hat{{\bf o}}_{id,1} {\rm \} }$. The backward track is a random trajectory that is expected to be similar to the correct forward track. The confidence of a track segments is defined as the distance between these two track segments: ${\rm Conf}(\tau _{t}^{} |\tau _{t}^{r} )={\rm distance}(\tau _{t}^{} ,\tau _{t}^{r} )$. We use the Euclidean distance between the initial point and the end point of the validation trajectory as ${\rm distance}(\tau _{t}^{} ,\tau _{t}^{r} )=||{\bf o}_{id,1:t} -{\bf o}_{id,t:1} ||$.

The MCMOT algorithm is summarized in the followings: 
\begin{algorithm}
  \SetKwInOut{Input}{Input}
  \SetKwInOut{Output}{Output}
  \Input{Motion model, entry model}
  \Output{Confident track segments}
  \textbf{Step 1.} Calculate the posterior $P{\rm (}{\bf x}_{t}^{} |{\bf z}_{1:t}^{t} {\rm )}$ \\
  \textbf{Step 2.} Generate track segments \\
  \textbf{Step 3.} Detect changing points for all track segments \\ 
  \textbf{Step 4.} Do forward-backward validation for the track segments with detected changing points \\
  \textbf{Step 5.} Generate resulting trajectories by combining the track segments
  \caption{MCMOT using CPD}
\end{algorithm}

\section{Experiment Results}

We describe the details about MCMOT experiment setting, and demonstrate the performance of MCMOT compared to the state-of-the-art methods in challenging video sequences.

\subsection{Implementation Details}

To build global and local object detector, we use publicly available sixteen-layer VGG-Net \cite{Ref19} and ResNet \cite{Ref29} which are pre-trained on an ImageNet classification dataset. We fine-tune an initial model using ImageNet Challenge Detection dataset (ImageNet DET) with 280K iterations at a learning rate of 0.001. After 280K iterations, the learning rate is decreased by a factor of 10 for fine-tuning with 70K iteration. For region proposal generation, RPN \cite{Ref28} is employed because it is fast and provides accurate region proposals in end-to-end manner by sharing convolutional features. After building initial model, we perform domain-adaptation for each dataset by fine-tuning with similar step described beforehand. Changing point detection algorithms used a two-stage time-series learning algorithm \cite{Ref24} which is computationally effective and achieves high detection accuracy. We consider as change point when change point score is greater than change point threshold. Change point threshold is empirically set as 0.3. 

\subsection{Dataset}

There are a few benchmark datasets available for multi-class multi-object tracking \cite{Ref43}. Since they deal with only two or three classes, we used benchmark datasets, ImageNet VID \cite{Ref31} and MOT 2016 \cite{Ref32}, where the former has 30 object classes and the latter is an up-to-date multiple object tracking benchmark. We compare its performance with state-of-the-arts on the ImageNet VID and MOT Benchmark 2016.

\subsubsection{ImageNet VID}

We demonstrate our proposed algorithm using ImageNet object detection from video (VID) task dataset \cite{Ref31}. ImageNet VID task is originally used to evaluate performance of object detection from video. Nevertheless, this dataset can be used to evaluate MCMOT because this challenging dataset consists of the video sequences recorded with a moving camera in real-world scenes with 30 object categories and the number of targets in the scene is changing over time. Object categories in these scenes take on different viewpoints and are subject to various degrees of occlusions. To ease the comparison with other state-of-the-arts, the performance of MCMOT on this dataset is primarily measured by mean average precision (mAP) which is used in ImageNet VID Challenge \cite{Ref31}. We use the initial release of ImageNet VID dataset, which consists of three splits which are train, validation, and test.

\subsubsection{MOT Benchmark 2016}

We evaluate our tracking framework on the MOT Benchmark \cite{Ref32}. The MOT Benchmark is an up-to-date multiple object tracking benchmark. The MOT Benchmark collects some new challenging sequences and widely used video sequences in the MOT community. MOT 2016 consists of a total of 14 sequences in unconstrained environments filmed with both static and moving cameras. All the sequences contain only pedestrians. These challenging sequences are composed with various configurations such as different viewpoints, and different weather condition. Therefore, tracking algorithms which are tuned for specific scenario or scene could not perform well. We adopt the CLEAR MOT tracking metrics \cite{Ref23} using MOT Benchmark Development Kit \cite{Ref32} for the evaluation.

\begin{figure}[t!]
\centering
\includegraphics[width=0.85\textwidth]{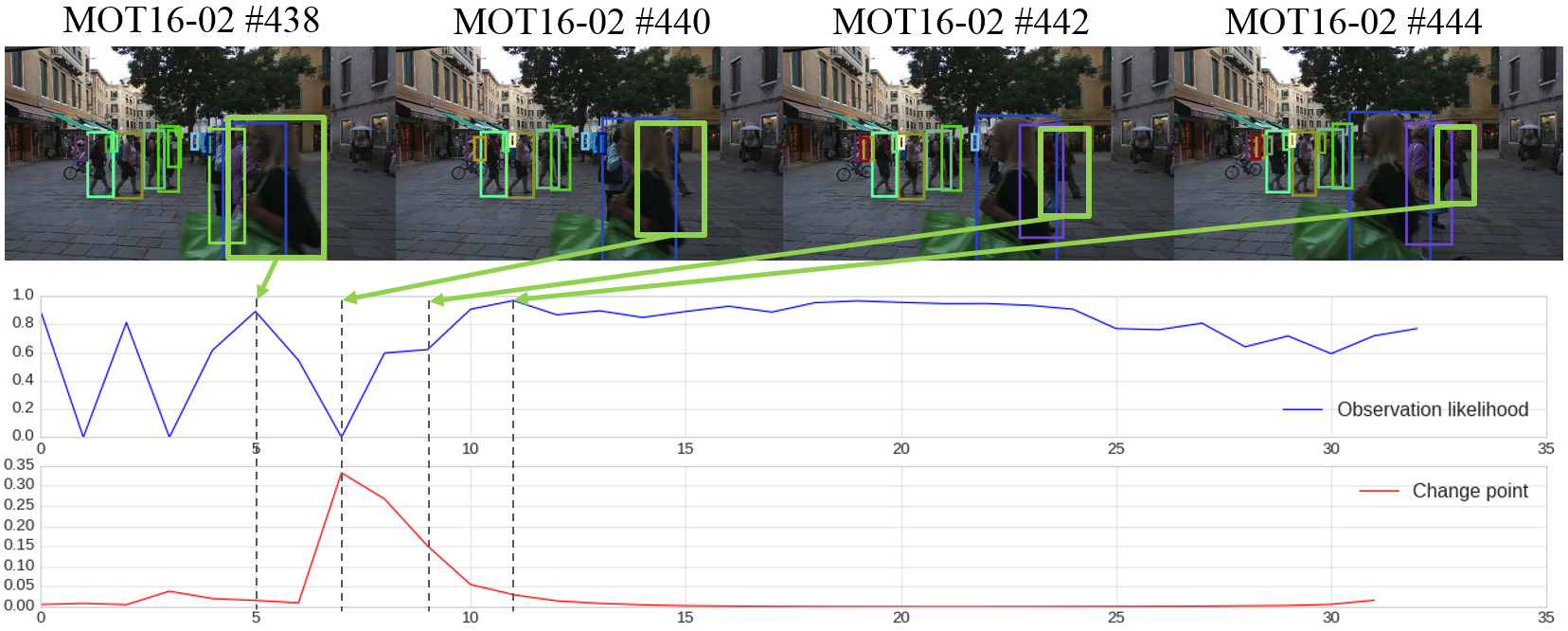} \\
(a) MOT16-02 sequence
\includegraphics[width=0.85\textwidth]{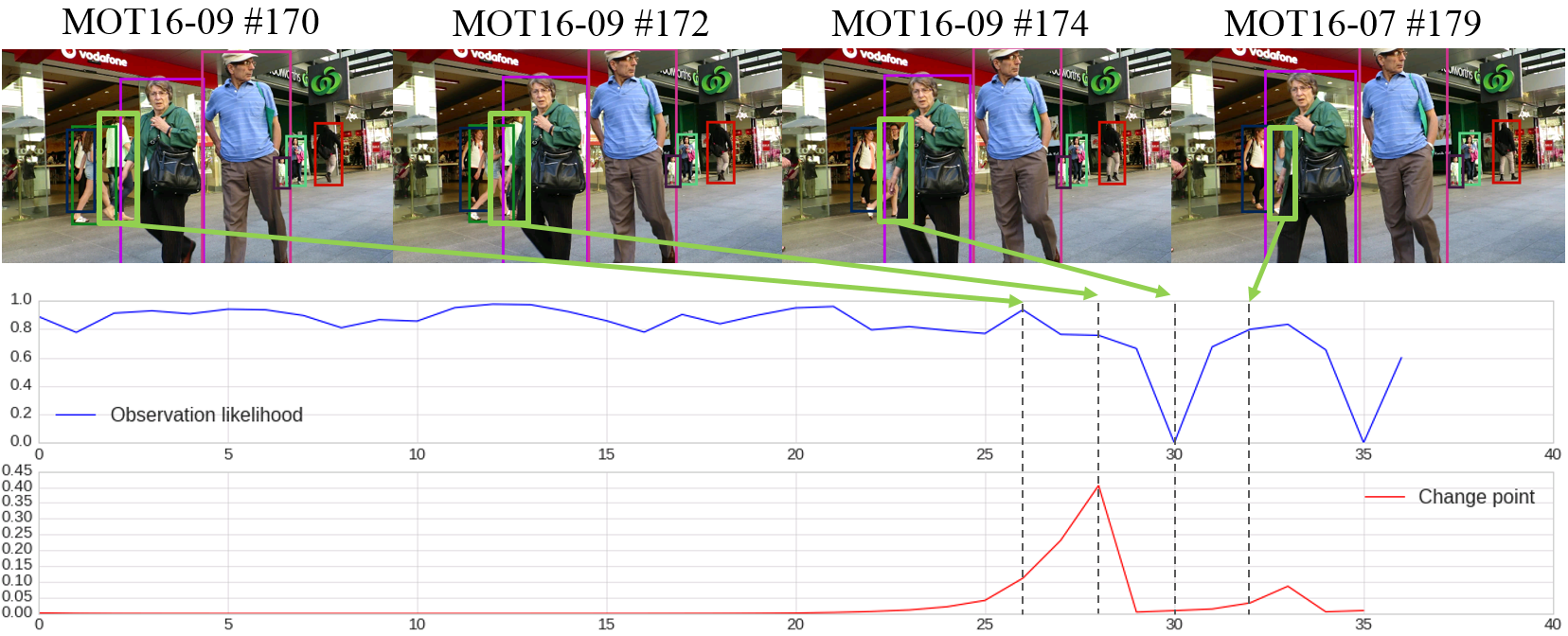} \\
(b) MOT16-09 sequence
\caption{Change points obtained from the segment in MOT16-02 and MOT16-09 sequence. A higher change point response indicates a higher uncertainty with high possibility of a drift occurrence. Notice that our method can effectively detect drifts in challenging situations.}
\label{fig:3}
\end{figure}

\subsection{MCMOT CPD Analysis}

In order to investigate the proposed MCMOT changing point detection component, we select two sequences, MOT16-02 and MOT16-09 from the MOT 2016 training set. For change point detection, we assign a change point if change point score is larger than 0.3. Fig.~\ref{fig:3} illustrates the observation likelihood and detected change point of the segment. A low likelihood or rapid change in likelihood is an important factor for detecting potential changing point. In the tracking result of MOT16-02 sequence in Fig.~\ref{fig:3}, unstable likelihood is observed until frame 438, where a motion-blurred half-body person moves. Tracking is drifted because occluded person appears at similar position with previous tracked point at frame 440. After several frames, the target is swapped to another person at frame 444. In this case, bounding boxes within drift area are unstable, which observed strong fluctuation of likelihood. Changing point detection algorithm produces high change point score at frame 440 by detecting this fluctuation. In the tracking result of MOT16-09 sequence in Fig.~\ref{fig:3} also illustrates similar situation explained before. As we can see, a possible track drift is implicitly handled by the change point detection method. 

\begin{table*}[!t]
\centering
\caption{Effect of different components on the ImageNet VID validation set}
\label{tab:1}
\scalebox{0.78}{
\tabcolsep=0.08cm
\begin{tabular}{l*{12}{c}}
\hline
\noalign{\smallskip}
 & aero & antelope & bear & bike & bird & bus & car & cattle & dog & cat & elephant \\
 \noalign{\smallskip}
\hline
\noalign{\smallskip}
Detection baseline & 84.6 & 75.8 & 77.2 & 57.2 & 60.8 & 84.6 & 62.4 & 66.3 & 57.7 & 62.3 & 74.0 \\
MCMOT CPD & 87.1 & 81.2 & 83.2 & 76.6 & 64.3 & 86.1 & 64.4 & 79.4 & 69.4 & 74.4 & 77.4 \\
MCMOT CPD FB & 86.3 & 83.4 & 88.2 & 78.9 & 65.9 & 90.6 & 66.3 & 81.5 & 72.1 & 76.8 & 82.4 \\
\noalign{\smallskip}
\hline
\noalign{\smallskip}
\hline
\noalign{\smallskip}
 & fox & g\_panda & hamster & horse & lion & lizard & monkey & m-bike & rabbit & r\_panda & sheep \\
 \noalign{\smallskip}
\hline
\noalign{\smallskip}
Detection baseline & 79.6 & 89.9 & 80.0 & 58.7 & 15.5 & 70.0 & 45.5 & 78.1 & 67.5 & 51.2 & 30.7 \\
MCMOT CPD & 87.3 & 90.2 & 85.3 & 63.3 & 31.7 & 74.8 & 52.6 & 86.9 & 74.7 & 75.2 & 30.5 \\
MCMOT CPD FB & 88.9 & 91.3 & 89.3 & 66.5 & 38.0 & 77.1 & 57.3 & 88.8 & 78.2 & 77.7 & 40.6 \\
\noalign{\smallskip}
\hline
\noalign{\smallskip}
\hline
\noalign{\smallskip}
 & snake & squirrel & tiger & train & turtle & boat & whale & zebra & \multicolumn{3}{c}{mean AP (\%)} \\
 \noalign{\smallskip}
\hline
\noalign{\smallskip}
Detection baseline & 50.7 & 29.0 & 79.5 & 71.5 & 68.9 & 77.0 & 57.9 & 77.9 & \multicolumn{3}{c}{64.7} \\
MCMOT CPD & 43.7 & 39.0 & 87.4 & 75.1 & 67.0 & 80.2 & 59.7 & 84.1 & \multicolumn{3}{c}{71.1} \\
MCMOT CPD FB & 50.3 & 44.3 & 91.8 & 78.2 & 75.1 & 81.7 & 63.1 & 85.2 & \multicolumn{3}{c}{\textbf{74.5}} \\
\noalign{\smallskip}
\hline
\end{tabular}
}
\label{table:1}
\end{table*}

\begin{table*}[!t]
\centering
\caption{Tracking performance comparison on the ImageNet VID validation set}
\label{tab:2}
\scalebox{0.78}{
\tabcolsep=0.08cm
\begin{tabular}{l*{12}{c}}
\hline
\noalign{\smallskip}
 & aero & antelope & bear & bike & bird & bus & car & cattle & dog & cat & elephant \\
 \noalign{\smallskip}
\hline
\noalign{\smallskip}
TCN \cite{Ref21} & 72.7 & 75.5 & 42.2 & 39.5 & 25.0 & 64.1 & 36.3 & 51.1 & 24.4 & 48.6 & 65.6 \\
ITLab VID-Inha & 78.5 & 68.5 & 76.5 & 61.4 & 43.1 & 72.9 & 61.6 & 61.1 & 52.2 & 56.6 & 74.0 \\
T-CNN \cite{Ref20} & 83.7 & 85.7 & 84.4 & 74.5 & 73.8 & 75.7 & 57.1 & 58.7 & 72.3 & 69.2 & 80.2 \\
\textbf{MCMOT (Ours)} & 86.3 & 83.4 & 88.2 & 78.9 & 65.9 & 90.6 & 66.3 & 81.5 & 72.1 & 76.8 & 82.4 \\
\noalign{\smallskip}
\hline
\noalign{\smallskip}
\hline
\noalign{\smallskip}
 & fox & g\_panda & hamster & horse & lion & lizard & monkey & m-bike & rabbit & r\_panda & sheep \\
 \noalign{\smallskip}
\hline
\noalign{\smallskip}
TCN \cite{Ref21} & 73.9 & 61.7 & 82.4 & 30.8 & 34.4 & 54.2 & 1.6 & 61.0 & 36.6 & 19.7 & 55.0 \\
ITLab VID-Inha & 72.5 & 85.5 & 67.5 & 64.7 & 5.7 & 54.3 & 34.7 & 77.6 & 53.5 & 40.8 & 34.3 \\
T-CNN \cite{Ref20} & 83.4 & 80.5 & 93.1 & 84.2 & 67.8 & 80.3 & 54.8 & 80.6 & 63.7 & 85.7 & 60.5 \\
\textbf{MCMOT (Ours)} & 88.9 & 91.3 & 89.3 & 66.5 & 38.0 & 77.1 & 57.3 & 88.8 & 78.2 & 77.7 & 40.6 \\
\noalign{\smallskip}
\hline
\noalign{\smallskip}
\hline
\noalign{\smallskip}
 & snake & squirrel & tiger & train & turtle & boat & whale & zebra & \multicolumn{3}{c}{mean AP (\%)} \\
 \noalign{\smallskip}
\hline
\noalign{\smallskip}
TCN \cite{Ref21} & 38.9 & 2.6 & 42.8 & 54.6 & 66.1 & 69.2 & 26.5 & 68.6 & \multicolumn{3}{c}{47.5} \\
ITLab VID-Inha & 18.1 & 23.4 & 69.6 & 53.4 & 61.6 & 78.0 & 33.2 & 77.7 & \multicolumn{3}{c}{57.1} \\
T-CNN \cite{Ref20} & 72.9 & 52.7 & 89.7 & 81.3 & 73.7 & 69.5 & 33.5 & 90.2 & \multicolumn{3}{c}{73.8} \\
\textbf{MCMOT (Ours)} & 50.3 & 44.3 & 91.8 & 78.2 & 75.1 & 81.7 & 63.1 & 85.2 & \multicolumn{3}{c}{\textbf{74.5}} \\
\noalign{\smallskip}
\hline
\end{tabular}
}
\label{table:2}
\end{table*}

\begin{figure}[t!]
\centering
\includegraphics[height=0.67in]{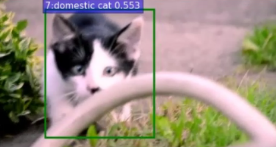}
\includegraphics[height=0.67in]{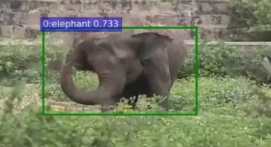}
\includegraphics[height=0.67in]{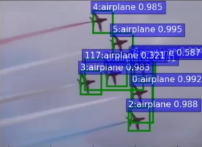}
\includegraphics[height=0.67in]{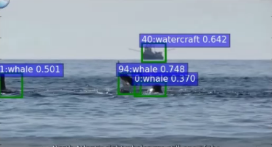}\\
\includegraphics[height=0.67in]{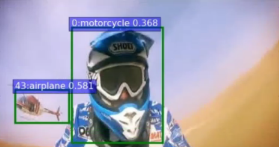}
\includegraphics[height=0.67in]{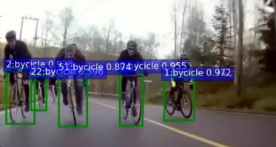}
\includegraphics[height=0.67in]{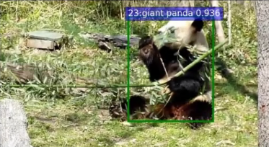}
\includegraphics[height=0.67in]{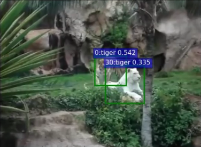}\\
\includegraphics[height=0.67in]{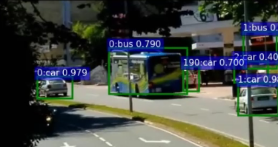}
\includegraphics[height=0.67in]{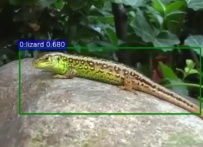}
\includegraphics[height=0.67in]{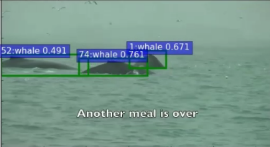}
\includegraphics[height=0.67in]{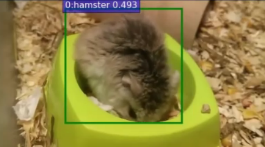}
\caption{MCMOT tracking results on the validation sequences in the ImageNet VID dataset. Each bounding box is labeled with the identity, the predicted class, and the confidence score of the segment. Viewing digitally with zoom is recommended.}
\label{fig:4}
\end{figure}

\subsection{ImageNet VID Evaluation}

Since the official ImageNet Challenge test server is primarily used for annual competition and has limited number of usage, we evaluate the performance of the proposed method on the validation set instead of the test set as a practical convention \cite{Ref20} for ImageNet VID task. For the ImageNet VID train/validation experiment, all the training and testing images are scaled by 600 pixel to be the length of image's shortest side. This value was selected so that VGG16 or ResNet fits in GPU memory during fine-tuning \cite{Ref28}.

Table~\ref{table:1} shows the effect of different components of MCMOT. Each method is distinguished in terms of MCMOT with CPD algorithm (MCMOT CPD), and MCMOT using CPD with forward-backward validation (MCMOT CPD FB). In the following evaluations, we filter out segments that have an average observation score lower than 0.3. As shown in the table~\ref{table:1}, significant improvement can be achieved with 9.8\% from detection baseline by adapting MCMOT CPD, and reached to 71.1\%. After the adaptation of the FB validation, an overall 74.5\% mAP was achieved on the ImageNet VID validation set. Table~\ref{table:2} summarizes the evaluation accuracy of MCMOT and the comparison with the other state-of-the-art algorithms on the whole 281 validation video sequences. Our MCMOT is achieved overall 74.5\% mAP on the ImageNet VID validation set, which is higher than state-of-the-art methods such as T-CNN \cite{Ref20}. This result is mainly due to the MCMOT approach of constructing a highly accurate segments by using CPD. As shown in Fig.~\ref{fig:4}, unlimited number of classes are successfully tracked with high localization accuracy using MCMOT.

\begin{table*}[!t]
\centering
\caption{Tracking performances comparison on the MOT benchmark 2016 (results on 7/14/2016). The symbol $\mathrm{\uparrow}$  denotes higher scores indicate better performance. The symbol $\mathrm{\downarrow}$ means lower scores indicate better performance}
\label{tab:3}
\scalebox{0.78}{
\tabcolsep=0.08cm
\begin{tabular}{l*{11}{c}}
\hline
\noalign{\smallskip}
Method & MOTA$\mathrm{\uparrow}$ & MOTP$\mathrm{\uparrow}$ & FAF$\mathrm{\downarrow}$ & MT$\mathrm{\uparrow}$ & ML$\mathrm{\downarrow}$ & FP$\mathrm{\downarrow}$ & FN$\mathrm{\downarrow}$ & ID Sw$\mathrm{\downarrow}$ & Frag$\mathrm{\downarrow}$ & Hz$\mathrm{\uparrow}$ \\
 \noalign{\smallskip}
\hline
\noalign{\smallskip}
GRIM & -14.5\% & 73.0\% & 10.0 & 9.9\% & 49.5\% & 59,040 & 147,908 & 1,869 & 2,454 & 10.0 \\
JPDA\_m \cite{Ref41} & 26.2\% & 76.3\% & 0.6 & 4.1\% & 67.5\% & 3,689 & 130,549 & 365 & 638 & 22.2 \\
SMOT \cite{Ref40} & 29.7\% & 75.2\% & 2.9 & 5.3\% & 47.7\% & 17,426 & 107,552 & 3,108 & 4,483 & 0.2 \\
DP\_NMS \cite{Ref39} & 32.2\% & 76.4\% & \textbf{0.2} & 5.4\% & 62.1\% & \textbf{1,123} & 121,579 & 972 & 944 & \textbf{212.6} \\
CEM \cite{Ref38} & 33.2\% & 75.8\% & 1.2 & 7.8\% & 54.4\% & 6,837 & 114,322 & 642 & 731 & 0.3 \\
TBD \cite{Ref37} & 33.7\% & 76.5\% & 1.0 & 7.2\% & 54.2\% & 5,804 & 112,587 & 2,418 & 2,252 & 1.3 \\
LINF1 & 41.0\% & 74.8\% & 1.3 & 11.6\% & 51.3\% & 7,896 & 99,224 & 430 & 963 & 1.1 \\
olCF & 43.2\% & 74.3\% & 1.1 & 11.3\% & 48.5\% & 6,651 & 96,515 & 381 & 1,404 & 0.4 \\
NOMT \cite{Ref22} & 46.4\% & 76.6\% & 1.6 & 18.3\% & 41.4\% & 9,753 & 87,565 & 359 & \textbf{504} & 2.6 \\
AMPL & 50.9\% & 77.0\% & 0.5 & 16.7\% & 40.8\% & 3,229 & 86,123 & \textbf{196} & 639 & 1.5 \\
NOMTwSDP16 \cite{Ref22} & 62.2\% & \textbf{79.6\%} & 0.9 & \textbf{32.5\%} & 31.1\% & 5,119 & 63,352 & 406 & 642 & 3.1 \\
\noalign{\smallskip}
\textbf{MCMOT\_HDM (Ours)} & \textbf{62.4\%} & 78.3\% & 1.7 & 31.5\% & \textbf{24.2\%} & 9,855 & \textbf{57,257} & 1,394 & 1,318 & 34.9 \\
\noalign{\smallskip}
\hline
\end{tabular}
}
\label{table:3}
\end{table*}

\begin{figure}[t!]
\centering
\hspace{0.15in} MOT16-03 \#100 \hspace{0.8in} MOT16-03 \#200 \hspace{0.5in} MOT16-06 \#100 \\
\includegraphics[height=0.94in]{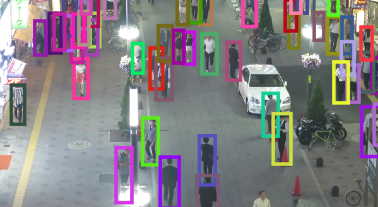}
\includegraphics[height=0.94in]{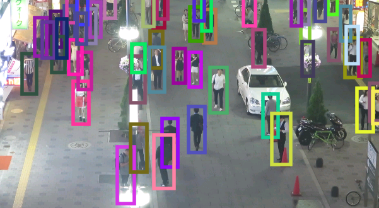}
\includegraphics[height=0.94in]{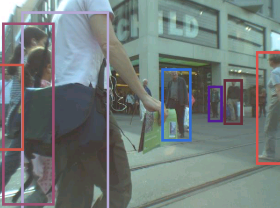} \\
MOT16-06 \#200 \hspace{0.5in} MOT16-07 \#100 \hspace{0.7in} MOT16-07 \#200 ~~~~~~ \\
\includegraphics[height=0.94in]{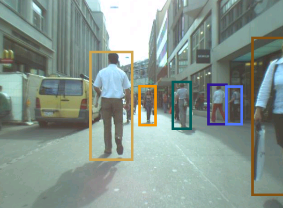}
\includegraphics[height=0.94in]{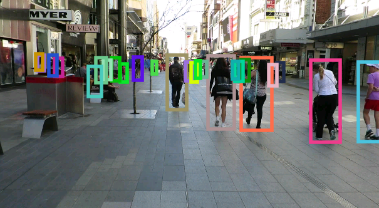}
\includegraphics[height=0.94in]{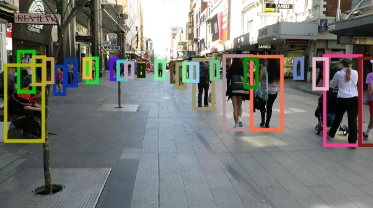} \\
MOT16-08 \#100 \hspace{0.6in} MOT16-08 \#200 \hspace{0.6in} MOT16-12 \#100 \\
\includegraphics[height=0.86in]{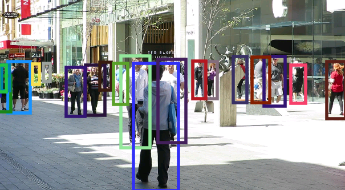}
\includegraphics[height=0.86in]{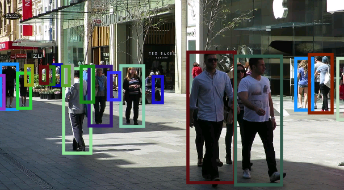}
\includegraphics[height=0.86in]{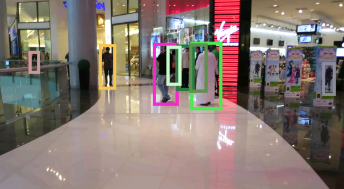} \\
MOT16-12 \#200 \hspace{0.6in} MOT16-14 \#100 \hspace{0.6in} MOT16-14 \#200 \\
\includegraphics[height=0.86in]{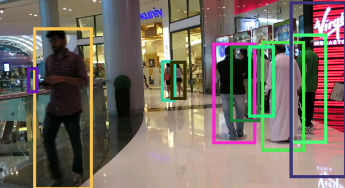}
\includegraphics[height=0.86in]{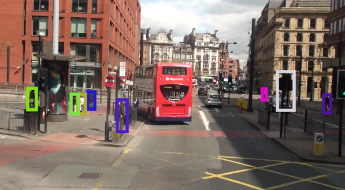}
\includegraphics[height=0.86in]{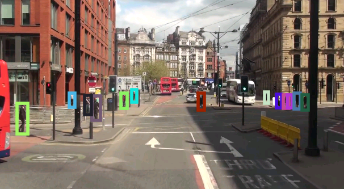}
\caption{MCMOT tracking results on the test sequences in the MOT Benchmark 2016. Each frame is sampled every 100 frames (these are not curated). The color of the boxes represents the identity of the targets. The figure is best shown in color.}
\label{fig:5}
\end{figure}

\subsection{MOT Benchmark 2016 Evaluation}

We evaluate MCMOT on the MOT Challenge 2016 benchmark to compare our approach with other state-of-the-art algorithms. For the MOT 2016 experiment, all the training and testing images are scaled by 800 pixel to be the length of image's shortest side. This larger value is selected because pedestrian bounding box size is smaller than ImageNet VID. In MCMOT, we also implement hierarchical data model (HDM) \cite{HDM} which is CNN based object detector. The timing excludes detection time.

Table~\ref{table:3} summarizes the evaluation metrics of MCMOT and the other state-of-the-arts on the test video sequences. Fig.~\ref{fig:5} visualizes examples of MCMOT tracking results on the test sequences. As shown in the table~\ref{table:3}, MCMOT outperforms the previously published state-of-the-art methods on overall performance evaluation metric which is called \textit{multi object tracking accuracy} (MOTA). We also achieved much smaller numbers of \textit{mostly lost targets} (ML) by a significant margin. Even though our method outperforms most of the metrics, \textit{tracker speed in frames per second} (HZ) is faster than other tracking methods. This is thanks to the simple MCMC tracking structure with entity status transition, and selective FB error validation with CPD, which is boosted tracking speed on a multi-object tracking task. However, high \textit{identity switch} (IDS) and high \textit{fragmentation} (FRAG) are observed because of the lack of identity mapping between track segments. More importantly, MCMOT achieves state-of-the-art performance in two different datasets, we demonstrate the general multi-class multi-obejct tracking applicability to any kind of situation with unlimited number of classes.

\section{Conclusion}
This paper presented a novel multi-class multi-object tracking framework. The framework surpassed the performance of state-of-the-art results on ImageNet VID and MOT benchmark 2016. MCMOT that cunducted unlimited object class association based on detection responses. The CPD model was used to observe abrupt or abnormal changes due to a drift. The ensemble of KLT based motion detector and CNN based object detector was employed to compute the likelihoods. A future research direction is to deal with the optimization problem of MCMOT structure and identity mapping problem between track segments.

\subsubsection{Acknowledgements} This work was supported by an Inha University research grant. A GPU used in this research was generously donated by NVIDIA Corporation.

\clearpage

\bibliographystyle{splncs03}


\end{document}